\documentclass[11pt]{article}
\usepackage{acl2016}
\usepackage{times}
\usepackage{url}
\usepackage{latexsym}
\usepackage{CJKutf8}
\usepackage{helvet}
\usepackage{courier}
\usepackage{multirow}
\usepackage{float}
\usepackage{stfloats}
\usepackage{amssymb,amsmath,mathrsfs,graphicx,epsfig,epstopdf}
\input zhwinfonts.tex
\usepackage{graphicx}
\usepackage{subfigure}

\begin{document}
\title{Can Machine Generate Traditional Chinese Poetry? A Feigenbaum Test}
\author{
Qixin Wang$^{1,4}$\footnotemark[1], Tianyi Luo$^{1,3}$\footnotemark[1], Dong Wang$^{1,2}$\footnotemark[2]\\
  $^1$CSLT, RIIT, Tsinghua University, China \\
  $^2$Tsinghua National Lab for Information Science and Technology, Beijing, China\\
  $^3$Huilan Limited, Beijing, China \\
  $^4$CIST, Beijing University of Posts and Telecommunications, China \\
  {\tt \{wqx, lty\}@cslt.riit.tsinghua.edu.cn} \\
  {\tt wangdong99@mails.tsinghua.edu.cn}
  \\
  }
\maketitle
\footnotetext[1]{The two authors contributed equally.}
\footnotetext[2]{Corresponding author: Dong Wang; RM 1-303, FIT BLDG, Tsinghua University, Beijing (100084), P.R. China.}
\begin{abstract}
  Recent progress in neural learning demonstrated that
  machines can do well in regularized tasks, e.g., the game of Go. However, artistic activities such as poem generation are still widely regarded
  as human's special capability. In this paper, we demonstrate that a simple neural model can imitate human in some tasks of art generation. We particularly
  focus on traditional Chinese poetry, and show that machines can do as well as many contemporary poets and weakly pass the Feigenbaum Test, a variant of Turing test in professional domains.

  Our method is based on an attention-based recurrent neural network, which accepts a set of keywords as the theme and generates poems by looking at each keyword during the generation. A number of techniques are
  proposed to improve the model, including
  character vector initialization,  attention to input and hybrid-style training. Compared to existing poetry generation methods, our model can generate much more theme-consistent and semantic-rich poems.

\end{abstract}

\section{Introduction}

The classical Chinese poetry is a special cultural heritage with over $2,000$ years of history and is still fascinating many contemporary poets.
In history, Chinese poetry flourished in different genres at different time, including Tang poetry, Song iambics and Yuan songs. Different genres possess their own specific structural, rhythmical and tonal patterns. The structural pattern regulates how many lines and how many characters per line; the rhythmical pattern requires that the last characters of certain lines hold the same or similar vowels; and the tonal pattern requires characters in particular positions hold particular tones, i.e., `Ping' (level tone), or `Ze' (downward tone). A good poem should follow all these pattern regulations (in a descendant order of priority), and has to express a consistent theme as well as a unique emotion. For this reason, it is widely admitted that traditional Chinese poetry generation is highly difficult and can be only performed by a very few knowledged people.

Among all the genres of traditional Chinese poetry, perhaps the most popular is the quatrain, a special style with a strict structure (four lines with five or seven characters per line), a regulated rhythmical form (the last characters in the second and fourth lines must follow the same rhythm), and a required tonal pattern (tones of characters in some positions should satisfy some pre-defined regulations). This genre of poems flourished mostly in Tang Dynasty, so often called `Tang poem'.  An example of quatrain written by Lun Lu, a famous poet in Tang Dynasty~\cite{wang2002book}, is shown in Table~\ref{tab:tang1}.

\begin{table}[!htb]
\begin{center}
\begin{tabular}{|c|c|c|}
\hline
\begin{CJK*}{UTF8}{gbsn}
塞下曲
\end{CJK*}
\\
Frontier Songs\\
\begin{CJK*}{UTF8}{gbsn}
月黑雁飞\textbf{高}，(*\,Z\,Z\,P\,P)
\end{CJK*}
\\
The wild goose flew high to \\
the moon shaded by the cloud,\\
\begin{CJK*}{UTF8}{gbsn}
单于夜遁\textbf{逃}。(P\,P\,Z\,Z\,P)
\end{CJK*}
\\
With the dark night's cover \\
escaped the invaders crowd, \\
\begin{CJK*}{UTF8}{gbsn}
欲将轻骑逐，(*\,P\,P\,Z\,Z)
\end{CJK*}
\\
I was about to hunt after \\
them with my cavalry,\\
\begin{CJK*}{UTF8}{gbsn}
大雪满弓\textbf{刀}。(*\,Z\,Z\,P\,P)
\end{CJK*}
\\
The snow already covered \\
 our bows and swords.\\
\hline
\end{tabular}
\end{center}
\caption{An example of a quatrain. The rhyming characters are in boldface, and the tonal pattern is shown at the end of each line, where `P' indicates level tone and `Z' indicates downward tone, and `*' indicates the tone can be either.}
\label{tab:tang1}
\end{table}

Due to the stringent restriction in rhythm and tone, it is not trivial to create a fully rule-compliant
quatrain. More importantly, besides such strict regulations, a good quatrain should also read fluently, hold a consistent theme, and express a unique affection. This is like
dancing in fetlers, hence very difficult and can be performed only by knowledged people with long-time training.

We are interested in machine poetry generation, not only because of its practical value in entertainment and education, but also because it demonstrates an important aspect of
artificial intelligence: the creativity of machines in art generation. We hold the belief that poetry generation (and other artistic activities) is a pragmatic process
and can be largely learned from past experience.
In this paper, we focus on traditional Chinese poetry generation, and demonstrate that machines can do it as well as
many human poets.

There have been some attempts in this direction, e.g., by machine translation models~\cite{he2012generating} and recurrent neural networks (RNN)~\cite{zhang2014chinese}. These methods can generate traditional Chinese poems with different levels of quality, and can be used to assist people in poem generation.
However, none of them can generate poems that are fluent and consistent enough, not to mention innovation.

In this paper, we propose a simple neural approach to traditional Chinese poetry generation based on the attention-based Gated Recurrent Unit (GRU) model. Specifically, we follow the sequence-to-sequence learning architecture that uses a GRU~\cite{cho2014learning} to encode a set of keywords as the theme, and another GRU to generate quatrains
character by character, where the keywords are looked back during the entire generation process. By this approach, the generation is regularized by the keywords so a global theme is assured. By enriching the set of keywords,
the generation tends to be more `innovative', resulting in more diverse poems. Our experiments demonstrated that the new approach can generate traditional Chinese poems pretty well and even pass the Feigenbaum Test.

\section{Related Work}

A multitude of methods have been proposed for poem automatic generation. The first approach is based on rules and templates. For example, \newcite{tosa2009hitch} and \newcite{wu2009new} employed a phrase search approach for Japanese poem generation, and~\newcite{netzer2009gaiku} proposed an approach based on word association norms. \newcite{oliveira2009automatic} and~\newcite{oliveira2012poetryme} used semantic and grammar templates for Spanish poem generation.

The second approach is based on various genetic algorithms. For example, \newcite{zhou2010genetic} proposed to use a stochastic search algorithm to obtain the best matched sentences. The search algorithm is based on four standards proposed by~\newcite{manurung2012using}: fluency, meaningful, poetic, and coherent.

The third approach to poem generation involves various statistical machine translation (SMT) methods. This approach was used by~\newcite{jiang2008generating} to generate Chinese couplets, a special regulated verses with only two lines.  \newcite{he2012generating} extended this approach to Chinese quatrain generation, where each line of the poem is generated by translating the preceding line.

Another approach to poem generation is based on text summarization. For example, \newcite{yan2013poet} proposed a method that retrieves high-ranking candidates of sentences from a large poem corpus, and then re-arranges the candidates to generate rule-conformed new sentences.

More recently, deep learning methods gain much attention in poem generation. For example, \newcite{zhang2014chinese} proposed an RNN-based approach that was reported to work well in quatrain generation~\cite{zhang2014chinese}; however, the structure seems rather complicated (a CNN and two RNN components in total), preventing it from extending to other genres. Our model is a simple sequence-to-sequence structure, which is much simpler than the model proposed by~\cite{zhang2014chinese} and can be easily extended to more complex genres such as Sonnet and Haiku without modification.

Finally, \newcite{wangchinese} proposed an attention-based model for Song Iambics generation. However, their model performed
rather poor when was applied directly to quatrain generation, possibly because quatrains are more condensed and more individually unique than iambics. Our approach follows the attention-based strategy in~\cite{wangchinese}, but introduces several innovations. Firstly, the poems were generated through key words rather than the first sentence to provide more clear themes; Secondly, a single-word attention mechanism was used to improve the sensitivity to key words; Thirdly, a loop generation approach was proposed to improve the fluency and coherence of the attention-based model.

\section{Method}

In this section, we first present the attention-based Chinese poetry generation framework, and then describe the implementation of the encoder and decoder models that have been tailored for our task.

\subsection{Attention-based Chinese Poetry Generation}

The attention-based sequence-to-sequence model proposed by~\newcite{bahdanau2014neural} is a powerful framework for
sequence generation. Specifically, the input sequence is converted by an `encoder' to a sequence of hidden states to represent the semantic status at each position of the input, and these hidden states are used to regulate a `decoder' that generates the target sequence. The important mechanism of the attention-based model is that at each generation step, the most relevant input units are discovered by comparing the `current' status of the decoder with all the hidden states of the encoder, so that the generation is regulated by the fine structure of the input sequence.

\begin{figure}[!htb]
\centering
\epsfig{figure=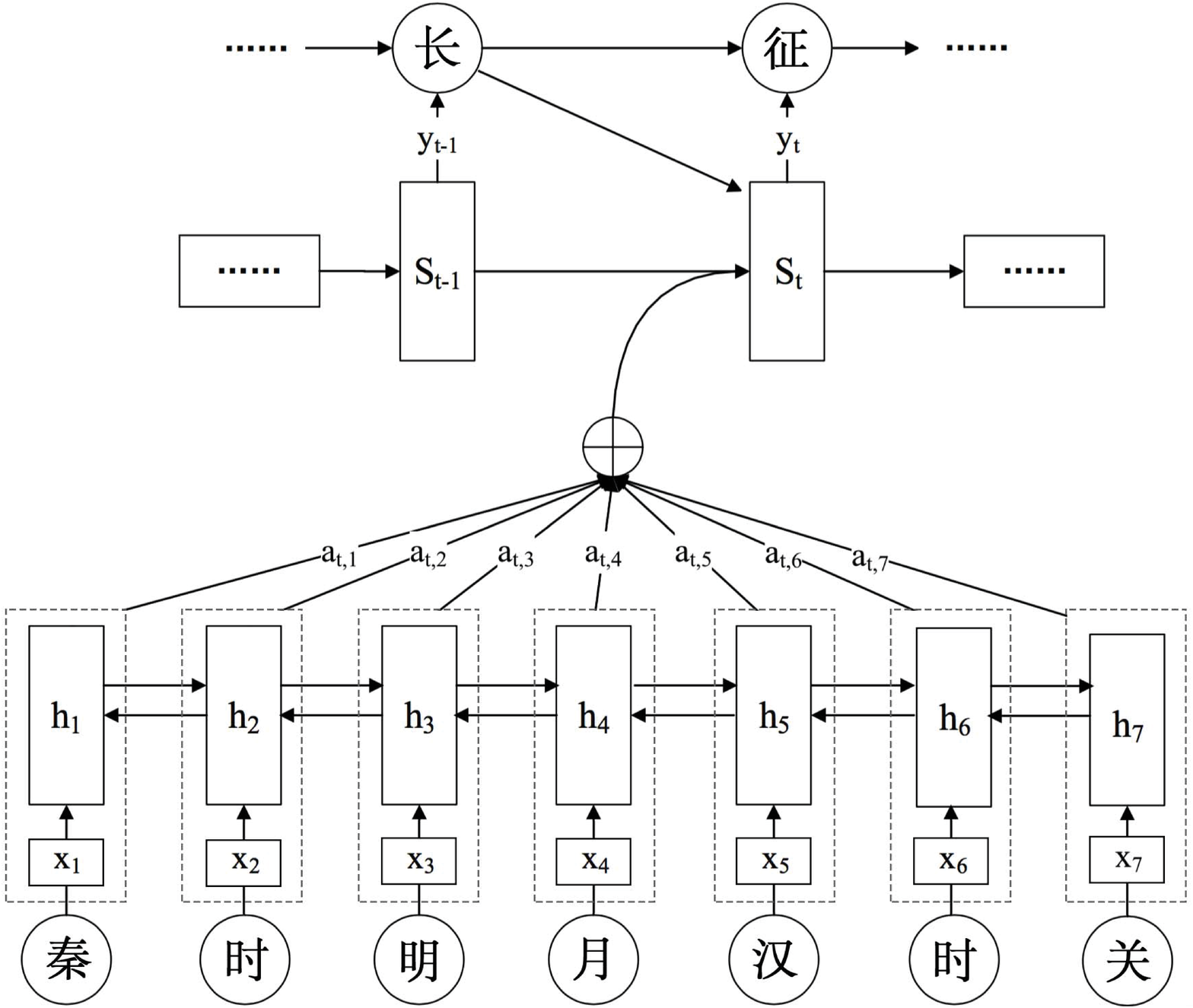,width=\linewidth}
\caption{The attention-based sequence-to-sequence learning framework for Chinese poetry generation.}
\label{fig:model}
\end{figure}

The entire framework of the attention-based model applied to Chinese poetry generation is illustrated in Figure~\ref{fig:model}. The encoder (a bi-directional GRU that will be discussed shortly) converts the input keywords, a character sequence denoted by $(x_1, x_2, ...)$, into a sequence of hidden states $(h_1,h_2,...)$.  The decoder then generates the whole poem character by character, denoted by $(y_1,y_2,...)$. At each step $t$, the prediction for the next character $y_{t}$ is based on the `current' status $s_t$ of the decoder as well as all the hidden states $(h_1,h_2,...,h_T)$ of the encoder. Each hidden state $h_i$ contributes to the generation
according to a relevance factor $\alpha_{t,i}$ that measures the similarity between $s_t$ and $h_i$.

\subsection{GRU-based Model Structure}

A potential problem of the RNN-based generation approach proposed by~\newcite{zhang2014chinese} is that the vanilla RNN used in their model tend to forget historical input quickly, leading to theme shift in generation. To
alleviate the problem, \newcite{zhang2014chinese} designed a composition strategy that generates only one line
at each time. This is certainly not satisfactory as it complicates the generation process.

In our model, the quick-forgetting problem is solved by using the GRU model. For encoder,
a bi-direction GRU is used to encode the input keywords, and for
the decoder, another GRU is used to conduct the generation. The GRU is powerful in remembering input and thus can
provide a strong memory for the theme, especially when combined with the attention mechanism.









\subsection{Model Training}

The goal of the model training is to let the predicted character sequence match the original poem. We chose the cross entropy between the distributions over Chinese characters given by the decoder and the ground truth (essentially in a one-hot form) as the objective function. To speed up the training, the minibatch stochastic gradient descent (SGD) algorithm was adopted. The gradient was computed sentence by sentence, and the AdaDelta algorithm was used to adjust the learning rate~\cite{zeiler2012adadelta}. Note that in the training phase, there are no keyword input, so we use the first line as the input to generate the entire poem.

\section{Implementation}

The basic attention model does not naturally work well for Chinese poetry generation. A particular problem is that every poem was created to express a special affection of the poet, so it tends to be `unique'. This means that most valid (and often great) expressions can not find sufficient occurrence in the training data. Another problem is that the theme may become vague towards the end of the generation, even with the attention mechanism. Several techniques are presented to improve the model.

\subsection{Character Vector Initialization}

Since the uniqueness of each poem, it is not simple to train the attention model from scratch, as many expressions are not statistically significant. This is a special form of data sparsity. A possible solution is to train the model in two steps: firstly learn the semantic representation of each character, possibly using a large external corpus, and then train the attention model with these pre-trained representations. By this approach, the attention model most focuses on possible expressions and hence is easier to train. In practice, we first derive character vectors using the word2vec tool\footnote{\url{https://code.google.com/archive/p/word2vec/}}, and then use these character vectors to initialize the word embedding matrix in the attention model. Since part of the model (embedding matrix) have been pre-trained, the problem of data sparsity can be largely alleviated.

\subsection{Input Reconstruction}

Poets tend to express their feelings following an implicit theme, instead of an explicit reiteration. We found this implicit theme is not easy for machines to understand and learn, leading to possible theme drift at run-time.
A simple solution is to force the model to reconstruct the input after it has generated the entire poem. More specifically, in the training phase, we use the first line of a training poem as the input, and based on this input to generate five lines sequentially: line $1$-$2$-$3$-$4$-$1$. The last generation step for line $1$ forces the model to keep the input in mind during the entire generation process, so leans how to focus on the theme.

\subsection{Input Vector Attention}

The popular configuration of the attention model attends on hidden states. Since hidden states
represent \emph{accumulated} semantic meaning, this attention is good to form a global theme.
However, as the semantic contents of individual keywords have been largely averaged, it is hard
 to generate diverse poems that are sensitive to each and different keywords.

We propose a multiple-attention solution that attends on both hidden states and input character
vectors, so that both accumulated and individual semantics are considered during the generation.
It has been found that this approach
is highly effective for generating diverse and novel poems: just given sufficient keywords, new poems can be generated
with high quality. Compared to other approaches such as noise injection or n-best inference,
this approach can generate unlimited alternatives without any quality sacrifice. Interestingly, our experiments show
that more keywords tend to generate more unexpected but highly impressive poems. Therefore, the
multiple-attention approach can be regard as an interesting way to promote innovation.

\subsection{Hybrid-style Training}

Traditional Chinese quatrains are categorized into 5-char quatrains and 7-char quatrains that involve five and seven characters per line, respectively. These two categories follow different regulations, but also share the same words
and similar semantics.
We propose a hybrid-style training that trains the two types of quatrains
using the same model, with a `type indicator' to notify the model which type
the present training sample is. In our study, the type indicators are derived from eigen vectors of a $200 \times 200$ dimensional random matrix. Each type of quatrain is assigned a fixed eigenvector as its type indicator.
The indicators are
provided as part of the input to the first hidden state of the decoder, and keep constant during the training.

\section{Experiments}

We describe the experimental settings and results in this section. Firstly the datasets used in the experiments are presented, and then we report the evaluation in three phases: (1) the first phase focuses on searching for optimal configurations for the attention model; (2) the second phase compares the attention model with other methods; (3) the third phase is the Feigenbaum Test.

\subsection{Datasets}
\label{sec:exp:data}

Two datasets are used to conduct the experiments. Firstly a Chinese quatrain corpus was collected from Internet. This corpus consists of $13,299$ $5$-char quatrains and $65,560$ $7$-char quatrains. As far as we know, this covers most of the quatrains that are retained today. We filters out some poems which contains 100\% low frequency words. Through corpus cleaning, a corpus which contains $9,195$ $5$-char quatrains and $49,162$ $7$-char quatrains was obtained. $9,000$ $5$-char and $49,000$ $7$-char quatrains are used to train the GRU model of the attention model and LSTM model of a comparative model based on RNN language models and the rest poems are used as the test datasets.

The second dataset was used to train and derive character vectors for attention model initialization. This dataset contains $284,899$ traditional Chinese poems in various genres, including Tang quatrains, Song iambics, Yuan Songs,
Ming and Qing poems. This large amount data ensures a stable learning for semantic content of most characters.

\subsection{Model Development}
\label{sec:exp:bleu}

In the first evaluation, we intend to find the best configurations for the proposed attention-based model.
The `Bilingual Evaluation Understudy' (BLEU)~\cite{papineni2002bleu} is used as the metric to determine which enhancement techniques are effective. BLEU was originally proposed to evaluate machine translation performance~\cite{papineni2002bleu}, and was used by~\newcite{zhang2014chinese} to evaluate quality of poem generation. We used BLEU as a cheap evaluation metric in the development phase to determine which design option to choose, without the costly human evaluation.

The method proposed by~\newcite{he2012generating} and employed by~\newcite{zhang2014chinese} was adopted to
obtain reference poems.
A slight difference is that the reference set was constructed for each \emph{input keyword}, instead of each sentence as in~\cite{zhang2014chinese}. This is because our attention model generates poems as an entire character sequence, while the vanilla RNN approach in~\cite{zhang2014chinese} does that sentence by sentence. Additionally, we used 1-gram and 2-grams in the BLEU computation, according to the fact that semantic meaning is mostly represented by single characters and some character pairs in traditional Chinese.

\begin{table}[!htb]
\begin{center}
\begin{tabular}{|l|c|c|}
\hline
Model                                                & \multicolumn{2}{|c|}{BLEU} \\
\cline{2-3}                                                  & 5-char & 7-char\\
\hline
Basic model                                          & 0.259  & 0.464\\
\,\,\,\,\,+ All poem training                        & 0.267  & 0.467\\
\,\,\,\,\,+ Input Reconstruction                     & 0.268  & 0.500\\
\,\,\,\,\,+ Input Vector Attention                   & 0.290  & 0.501\\
\,\,\,\,\,+ Hybrid training                          & {\bf0.330}  & {\bf0.630}\\
\hline
\end{tabular}
\end{center}
\caption{\label{tab:bleu} BLEU scores with various enhancement techniques. }
\end{table}

\begin{table*}[tp]
\begin{center}
\begin{tabular}{|c|c|c|c|c|c|c|c|c|c|c|c|c|c|}
\hline
Model       & \multicolumn{2}{|c|}{Compliance}       & \multicolumn{2}{|c|}{Fluency }               & \multicolumn{2}{|c|}{Consistence} &\multicolumn{2}{|c|}{Aesthesis}  & \multicolumn{2}{|c|}{Overall} \\ 
 \cline{2-11}    & char-5     & char-7   & char-5     & char-7      &   char-5   & char-7 &   char-5   & char-7 &   char-5   & char-7 \\
\hline
SMT    & 3.04     & 2.83    & 2.28     & 1.92      &   2.15    & 2.00 &   1.93    & 1.67 &2.35 &2.10 \\
\hline
LSTMLM       & 3.00     & 3.71   & 2.39      & 3.10  & 2.19 & 2.88 &   2.00    & 2.66 &2.39 &3.08\\
\hline
RNNPG   & 2.90     & 2.60  & 2.05        &    1.70    &   1.97   & 1.70 &   1.70    & 1.45& 2.15 &1.86\\
\hline
Attention   & {\bf3.44}     & {\bf3.73}  &  2.85          & 3.13    &   2.77   &2.98 &   2.38    & 2.87&2.86 &3.17\\
\hline
\hline
Human  & 3.33     & 3.54   &  {\bf3.37}         & {\bf3.33}    &   {\bf3.45}   &{\bf3.26} &   {\bf3.05}    & {\bf2.96}&{\bf3.30} &{\bf3.27}\\
\hline
\end{tabular}
\end{center}
\caption{\label{tab:bestresult1} Averaged ratings for Chinese quatrain generation with different methods. `char-5' and `char-7' represent 5-char and 7-char characters quatrains respectively in the evaluation. }
\end{table*}

Table~\ref{tab:bleu} presents the results. The baseline model is trained with character initialization
where the character vectors are trained using quatrains only. This is mostly the system in~\cite{wangchinese}.
Then we use the large corpus that involves
all traditional Chinese poems to enhance the character vectors, and the results demonstrated a noticeable performance improvement in fluency (from our human judgements) and a small improvement in BLEU ($2$nd row in Table~\ref{tab:bleu}). This is understandable since poems in different
genres use similar languages, so involving more training data helps infer more reliable semantic
content for each character.
Additionally, we observe that reconstructing the input during model training improves the model ($3$rd row).
This is probably due to the enhancement in theme consistence. What's more, attention to both input vectors and
hidden states leads to additional performance gains ($4$th row). Finally, the hybrid-style training
is employed to train a single model for the 5-char and 7-char quatrains. The BLEUs are tested on
5-char and 7-char quatrains respectively and the results are shown in the $5$-th row of Table~\ref{tab:bleu}.
Note that in the hybrid training, we stop the training before convergence in favor of a good BLEU.

From these results, we obtain the best configuration that involves character vector trained with extern
training data, input reconstruction, input vector attention and hybrid training.
In the reset of the paper, we will use this configuration to train the attention model (denoted by `Attention') and compare it with the comparative methods.

\subsection{Comparative Evaluation}
\label{sec:exp:human}

In the second phase, we compare the attention model (with the best configuration) and three comparative models: the SMT model proposed by~\newcite{he2012generating}, the vanilla RNN poem generaion (RNNPG) proposed by~\newcite{zhang2014chinese}, and an RNN language model (RNNLM) that can be regarded as a simplified version (One-direction LSTM RNN neural network without attention mechanism) of the attention model~\cite{mikolov2010recurrent}.

Following the work of~\newcite{zhang2014chinese}, we selected $30$ subjects (e.g., falling flower, stone bridge, etc.) in the Shixuehanying taxonomy~\cite{liu1735book} as $30$ themes. For each theme, several phrases belonging to the corresponding subject were selected as the input keywords. For the attention model, these keywords were used to generate the first line directly; For the other three models, however, the first line had to be constructed beforehand by an
external model. We chose the method provide by~\newcite{zhang2014chinese} to generate the first lines for the SMT, vanilla RNN and LSTMLM approaches. A $5$-char quatrain and a $7$-char quatrain were generated for each theme by the four methods, and were evaluated by experts.

For reference, some poems written by ancient poets were also involved in the evaluation. Note that to prevent the impact of prior knowledge of the experts, we deliberately chose the poems that were written by poets that are not very famous. The poems were chosen from~\cite{han2015book},~\cite{Kojiro1963book} and~\cite{chen2013book}; and a $5$-char quatrain and a $7$-char quatrain were selected for each theme.

The evaluation was conducted by experts based on the following four metrics, in the scale from $0$ to $5$:

\begin{itemize}
\item Compliance: if the poem satisfies the regulation on tones and rhymes;
\item Fluency: if the sentences read fluently and convey reasonable meaning;
\item Consistence: if the poem adheres to a single theme;
\item Aesthesis: if the poem stimulates any aesthetic feeling.
\end{itemize}

In the experiments, we invited $26$ experts to conduct a series of scoring evaluations\footnote{These experts are professors and their postgraduate students in the field of Chinese poetry research. Most of them are from the Chinese Academy of Social Sciences (CASS).}. These experts were asked to rate the generation of our model and three comparative approaches: SMT, LSTMLM, and RNNPG. The SMT-based approach is available online\footnote{http://duilian.msra.cn/jueju/} and we use this online service
to obtain the generation. For RNNPG, we invited the authors to conduct the generation for us. The LSTMLM approach was
implemented by ourselves, for which we used the GRU instead of the vanilla RNN to enhance long-distance memory, and used character vector initialization to improve model training.

Poems written by ancient poets are also
involved in the test. For each method (including human-written), a $5$-char quatrain and a $7$-char quatrain were
generated or selected for each of the $30$ themes, amounting to $300$ poems in total in the test. For each expert,  $80$ poems were randomly selected for evaluation.

Table~\ref{tab:bestresult1} presents the results. It can be seen that our model outperforms all the comparative  approaches in terms of all the four metrics. More interestingly, we find that the scores obtained by our model
are approaching to those obtained by human poets, especially with 7-char poems. This is highly encouraging and
indicates that our model can imitate human beings to a large extent, at least from the eyes of contemporary experts.

The second best approach is the LSTMLM approach. As we mentioned,
LSTMLM can be regarded as a simplified version of our attention model, and shares the same strength in LSTM-based
long-distance pattern learning and improved training strategy with character vector initialization. This
demonstrated that a simple neural model with little engineering effort may learn artistic activities pretty good.
Nevertheless, the comparative advantage of the attention model still demonstrated the importance of the
attention mechanism.

The RNNPG and the SMT approaches perform equally worse, particularly RNNPG.
A possible reason is that RNNPG requires an SMT model to enhance the performance but the
SMT model was not used in this test\footnote{The author of RNNPG~\cite{zhang2014chinese} could not find
the SMT model in the reproduction, unfortunately.}. In fact, even with the SMT model, RNNPG can hardly approach
to human as the attention model does, as shown in the original paper~\cite{zhang2014chinese}.
The SMT approach, with a bunch of unknown optimizations by the Microsoft colleagues, can deliver reasonable
quality, but the limitation of the model prevents it from approaching a human-level as our model does. The T-test results show that the difference between the attention LSTM model (ours) and the vanilla RNN and SMT are both significant ($p$ < 0.01), though the difference between the attention LSTM model and LSTMLM is weakly significant ($p$ = 0.03).

It is noticeable that the human ratings of human-written poems are lower than the ratings reported by ~\newcite{zhang2014chinese}. We are not sure the experts that Zhang and Lapata invited, but the experts in our experiments are truly professional and critical: most of them are top-level experts on classical Chinese poetry education and criticism, and some of them are winners of national competitors in classic Chinese poetry writing.

Additionally, we note that almost in all the evaluations, the human-written poems beat those generated by machines. On one hand, this indicates that human are still superior in artistic activities, and on the other hand, it demonstrates from another perspective that the participants of the evaluation are truly professional and can tell good or bad poems. Interestingly, in the metric of compliance, our attention model outperforms human. This is not surprising as computers can simply search vast candidate characters to ensure a rule-obeyed generation. In contrast, human artists put meaning and affection as the top priority, so sometimes break the rule.

Finally, we see that the quality of the $7$-char poems generated by our model is very close to that of the human-written poems. This should be interpreted in two perspectives: On one hand, it indicates that our generation is rather successful; On the other hand, we should pay attention that the poems we selected are from unfamous poets.
Our intention was to avoid biased rating caused by experts' prior knowledge on the poems, but this may have
limited the quality of the selected poems, although we have tried our best to choose good ones.

\subsection{Feigenbaum Test}

We design a Feigenbaum Test~\cite{feigenbaum2003some} to evaluate the quality of poems generated by our models.
Feigenbaum test (FT) can be regarded as an generalized Turing test (TT), the most well-known method
for evaluating AI systems. A particular shortcoming of TT is that it is only suitable for tasks involving
interactive conversions. However, there are many professional domains where no conversations are involved
but still require a simple method like TT to evaluate machines' intelligence. Feigenbaum Test follows the core idea of TT, but focuses on professional tasks that can be done only by domain experts. The basic idea of FT is
that an intelligent system in a professional domain should behave as a human expert, and the behavior can
not be \emph{distinguished from} human experts, when \emph{judged by} human experts in the same domain. We believe
that this is highly important when evaluating AI systems on artistic activities, for which mimicking the behavior of
human experts is an important indicator of its success.

In this section, we follow this idea and utilize FT to evaluate the poetry generation models. Specifically, we distributed the $30$ themes to some experts in traditional Chinese poem generation\footnote{These experts were nominated by professors in the field of traditional Chinese poetry research.}. We asked these experts to select one theme that they are most favor so that the quality can be ensured.

We received $144$ submissions. To ensure the quality of the submission, we generated the same number of poems by our
model and then asked a highly-professional expert in traditional Chinese poem criticism to give the first round filtering.
After the filtering, $83$ human-written poems ($57.6\%$) and $180$ computer-generated poems ($86.5\%$) were remained, respectively.
This indicates that human-generated poems are in a larger variance in quality, which is not very surprising
as the knowledge and skill of people tend to vary significantly.

The remained $263$ poems were distributed to $24$ experts for evaluation\footnote{These
experts again are mostly from CASS, and some of them attended the previous test but not all.}. The experts were
asked to answer two questions: (1) if a poem was generated by people; (2) quality of a poem, rated from $0$ to $5$.

\begin{figure} [htb]
\subfigure[] {
\label{fig:a}
\includegraphics[width=4cm]{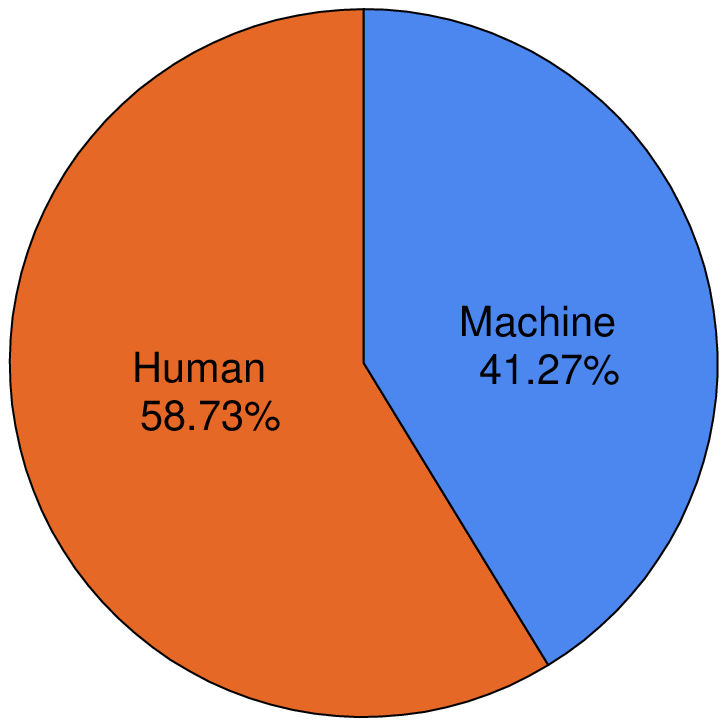}
}%
\subfigure[] {
\label{fig:b}
\includegraphics[width=4cm]{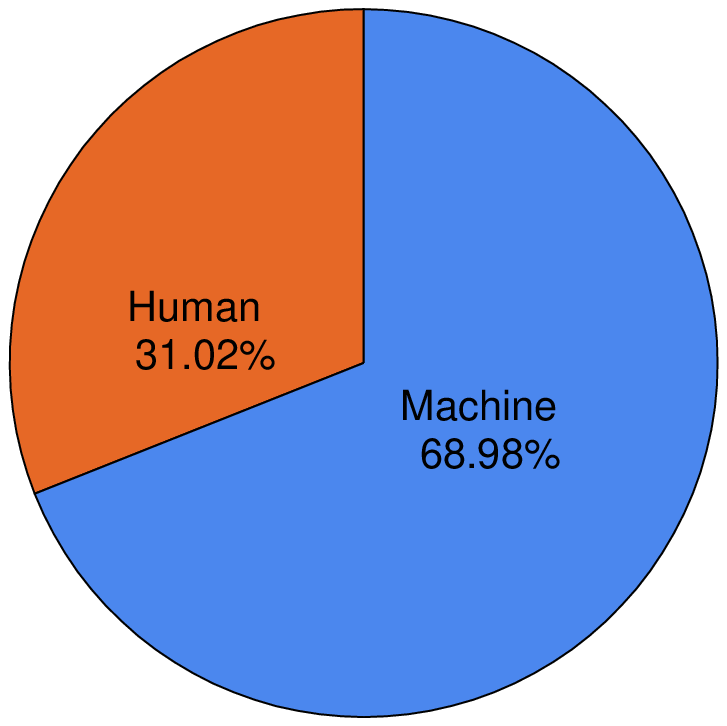}
}
\caption{Decision option for (a) human-written  (b) machine-generated poems.}
\label{fig:opt}
\end{figure}

\begin{figure} [htb]
\subfigure[] {
\label{fig:c}
\includegraphics[width=4cm]{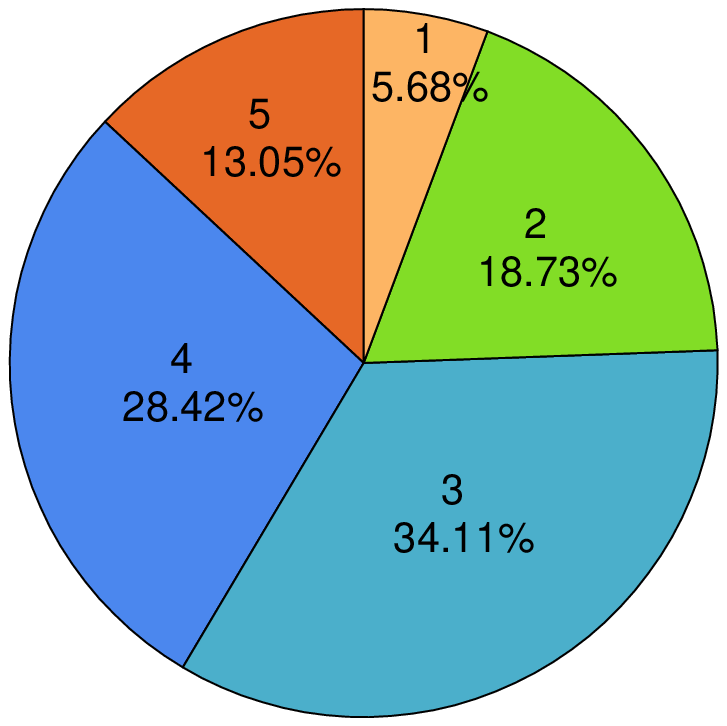}
}%
\subfigure[] {
\label{fig:d}
\includegraphics[width=4cm]{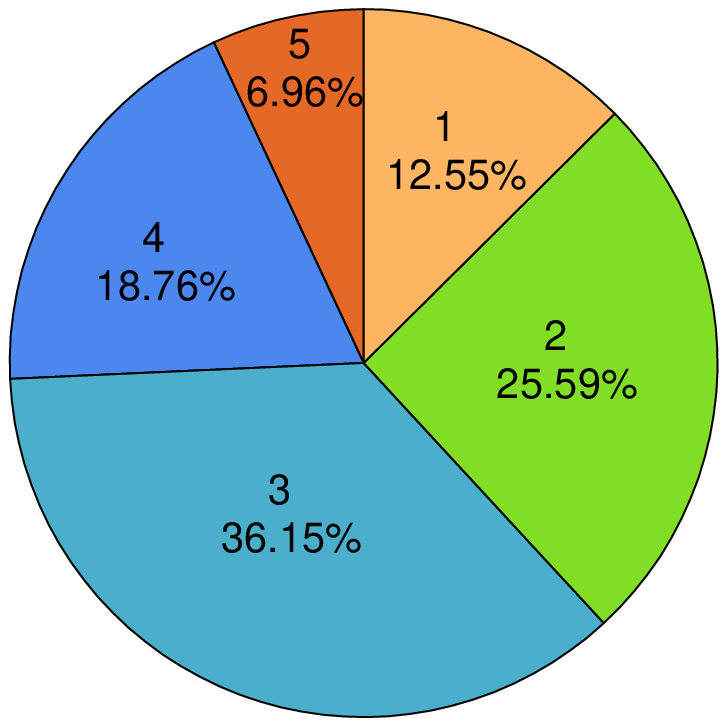}
}

\caption{Score distribution for (a) human-written (b) machine-generated poems.}
\label{fig:score}
\end{figure}

The results for the human-machine decision are presented in Figure~\ref{fig:opt}. For a clear
representation, the minor proportions of zero scores are omitted in the figure.
We observe that 41.27\% of the human-written poems were identified as machine-generated, and 31\% of the machine-generated poems were identified as human-written. This indicates that a large number of poems can not be correctly identified by people. According to the criterion of Turing Test(Actually, Feigenbaum Test can be regarded as domain specific "Turing Test"), our model has weakly passed\footnote{The criterion is to fool people in more than 30\% of the trials. Refer to \url{https://en.wikipedia.org/wiki/Turing_test}.}. The score distributions for human-written and machine-generated poems are presented in Figure~\ref{fig:score}. It can be seen that our model is still inferior to human in average. However, a large proportion (61.9\%)
of the machine-generated poems were scored equal to or more than $3$, which means that our model works pretty well, as human poets can only achieves $75.6\%$.
Interestingly, among the top-5 high-ranked poems, the machine takes the position $1$ and $2$, and among the top-10 high-ranked poems, the machine takes the position $1$, $2$ and $7$. This means that our model can generate very good poems, even better than human poets, although in general it is still beat by human.



\begin{figure}[!htb]
\centering
\epsfig{figure=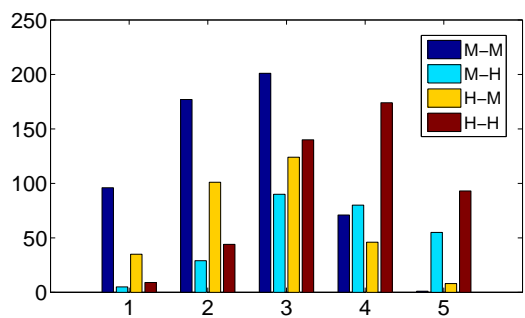,width=\linewidth}
\caption{Score distribution for the poems written by and identified as the two types of authors (human or machine). In the figure, `M-H' means poems generated by machine but identified as human-written.}
\label{fig:MMMHHMHH}
\end{figure}

A more detailed analysis is presented in Figure~\ref{fig:MMMHHMHH}, where the poems are categorized into four groups according to their `genuine' and `identified' authors (human or
machine). From the two pairs M-M vs. M-H and H-M vs. H-H, we observe that a poem tends to be rated high if the experts consider them as Human-written. This means that the identification is positively related to the score, and people still tend to recognize human writes better. This is also true anyway at present.



To have a better understanding of the decision process, we invited another $4$ experts to specify the metrics by which the human-machine identification was made for each poem. Multiple metrics can be selected. The proportions that each metric was selected are shown in Table~\ref{tab:dec-factor}. It can seen that the experts tend to regard fluency and aesthesis as the most important
factors in the decision. When evaluating human-written poems, it shows that a fluent poem tends to be identified correctly, while a poem without any aesthetic feeling tends to be recognized as machine-generated.

\begin{table}[!htb]
\begin{center}
\begin{tabular}{|l|c|c|c|c|c|}
\hline
        & Comp. & Flu. & Cons. & Aes.  \\
        \hline
M-M & 20.8\%    & {\bf68.0\%} & 51.2\% & 60.0\% \\
M-H & 22.9\%    & 51.4\% & {\bf54.3\%} & 51.4\%   \\
H-M & 32.9\%    & 51.8\% & 45.9\% & {\bf76.5\%}  \\
H-H & 10.7\%    & {\bf72.0\%} & 58.7\% & 61.3\%  \\
\hline
Overall  & 21.9\%    &  62.8\%  &  51.9\%   &  {\bf63.8\%}     \\
\hline
\end{tabular}
\end{center}
\caption{\label{tab:dec-factor} Percentage of each metric was chosen in the identification decision.
`M-H' means the category that machine-generated poems are identified as human-written.}
\end{table}



\subsection{Generation Example}

Finally we show a 7-char quatrain generated by the attention model. The theme of this poem is `crab-apple flower'.

\begin{table}[!htb]
\begin{center}
\begin{tabular}{|c|c|c|}
\hline
\begin{CJK*}{UTF8}{gbsn}
海棠花
\end{CJK*}
\\
Crab-apple Flower\\
\begin{CJK*}{UTF8}{gbsn}
红霞淡艳媚妆水，
\end{CJK*}
\\
Like the rosy afterglows with \\
light make-up being sexy,\\
\begin{CJK*}{UTF8}{gbsn}
万朵千峰映碧垂。
\end{CJK*}
\\
Among green leaves, thousands of crabapples\\
 blossoms make the branch droopy.\\
\begin{CJK*}{UTF8}{gbsn}
一夜东风吹雨过，
\end{CJK*}
\\
After a night of wind and shower,\\
\begin{CJK*}{UTF8}{gbsn}
满城春色在天辉。
\end{CJK*}
\\
With the bright sky, spring \\
is all over the city.\\
\hline
\end{tabular}
\end{center}
\caption{A quatrain example generated by the attention model.}
\label{tab:song}
\end{table}

\section{Conclusion}

This paper proposed an attention-based neural model for Chinese poetry generation.
Compared to existing methods, the new approach is simple in model structure, strong
in theme preservation, flexible to produce innovation, and easy to
be extended to other genres. Our experiments show that it can generate traditional
Chinese quatrains pretty well and weakly pass the Feigenbaum Test. A future work will
employ more generative models,
e.g. variational generative deep models, to achieve more
natural innovation. We also plan to extend the work to other genres of traditional Chinese poetry, e.g., Yuan songs.

\clearpage
\newpage

\bibliography{acl2016}
\bibliographystyle{acl2016}

\end{document}